\documentclass[letterpaper, 10 pt, conference]{ieeeconf}  % Comment this line out if you need a4paper

\IEEEoverridecommandlockouts                              % This command is only needed if 
\usepackage[utf8]{inputenc} % allow utf-8 input
\usepackage[T1]{fontenc}    % use 8-bit T1 fonts
\usepackage{hyperref}       % hyperlinks
\usepackage{url}            % simple URL typesetting
\usepackage{booktabs}       % professional-quality tables
\usepackage{amsfonts}       % blackboard math symbols
\usepackage{nicefrac}       % compact symbols for 1/2, etc.
\usepackage{microtype}      % microtypography

\usepackage{times}
\usepackage{graphicx} % more modern
\usepackage{caption}
\usepackage{subcaption}
\usepackage{wrapfig}
\usepackage{rotating}

\usepackage{algorithm}
\usepackage{algorithmic}

\usepackage{hyperref}

\usepackage{amsmath}
\usepackage{amssymb}

\usepackage[font=footnotesize]{caption,subcaption}
\usepackage{sidecap}

\usepackage{xspace}

\usepackage{color}

\usepackage{pifont}

\newcommand{\specialcell}[2][c]{%
  \begin{tabular}[#1]{@{}c@{}}#2\end{tabular}}
  
\newcommand{\ba}{\mathbf{a}}
\newcommand{\bA}{\mathbf{A}}
\newcommand{\bs}{\mathbf{s}}
\newcommand{\dataset}{\mathcal{D}}
\newcommand{\dsim}{\dataset^{\textsc{sim}}}
\newcommand{\dreal}{\dataset^{\textsc{rw}}}
\newcommand{\pith}{\pi_{\theta}}

\newcommand{\E}{\mathbb{E}}

\newcommand{\cmark}{\ding{51}}%
\newcommand{\xmark}{\ding{55}}%

\newcommand\rurl[1]{%
  \href{http://#1}{\nolinkurl{#1}}%
}

\title{\LARGE \bf
Generalization through Simulation:\\Integrating Simulated and Real Data into Deep Reinforcement Learning\\for Vision-Based Autonomous Flight
}

\author{Katie Kang$^{*}$, Suneel Belkhale$^{*}$, Gregory Kahn${^*}$, Pieter Abbeel, Sergey Levine\\% <-this % stops a space
Berkeley AI Research (BAIR), University of California, Berkeley%
\thanks{$^{*}$These authors contributed equally to this work.}
}

\begin{document}

\maketitle
\thispagestyle{empty}
\pagestyle{empty}

%%%%%%%%%%%%%%%%%%%%%%%%%%%%%%%%%%%%%%%%%%%%%%%%%%%%%%%%%%%%%%%%%%%%%%%%%%%%%%%%
\begin{abstract}
Deep reinforcement learning provides a promising approach for vision-based control of real-world robots. However, the generalization of such models depends critically on the quantity and variety of data available for training. This data can be difficult to obtain for some types of robotic systems, such as fragile, small-scale quadrotors. Simulated rendering and physics can provide for much larger datasets, but such data is inherently of lower quality: many of the phenomena that make the real-world autonomous flight problem challenging, such as complex physics and air currents, are modeled poorly or not at all, and the systematic differences between simulation and the real world are typically impossible to eliminate. In this work, we investigate how data from both simulation and the real world can be combined in a hybrid deep reinforcement learning algorithm. Our method uses real-world data to learn about the dynamics of the system, and simulated data to learn a generalizable perception system that can enable the robot to avoid collisions using only a monocular camera. We demonstrate our approach on a real-world nano aerial vehicle collision avoidance task, showing that with only an hour of real-world data, the quadrotor can avoid collisions in new environments with various lighting conditions and geometry. Code, instructions for building the aerial vehicles, and videos of the experiments can be found at \rurl{github.com/gkahn13/GtS}
\end{abstract}

%%%%%%%%%%%%%%%%%%%%%%%%%%%%%%%%%%%%%%%%%%%%%%%%%%%%%%%%%%%%%%%%%%%%%%%%%%%%%%%%
\section{Introduction}

\begin{figure}
    \centering
    \includegraphics[width=0.8\columnwidth]{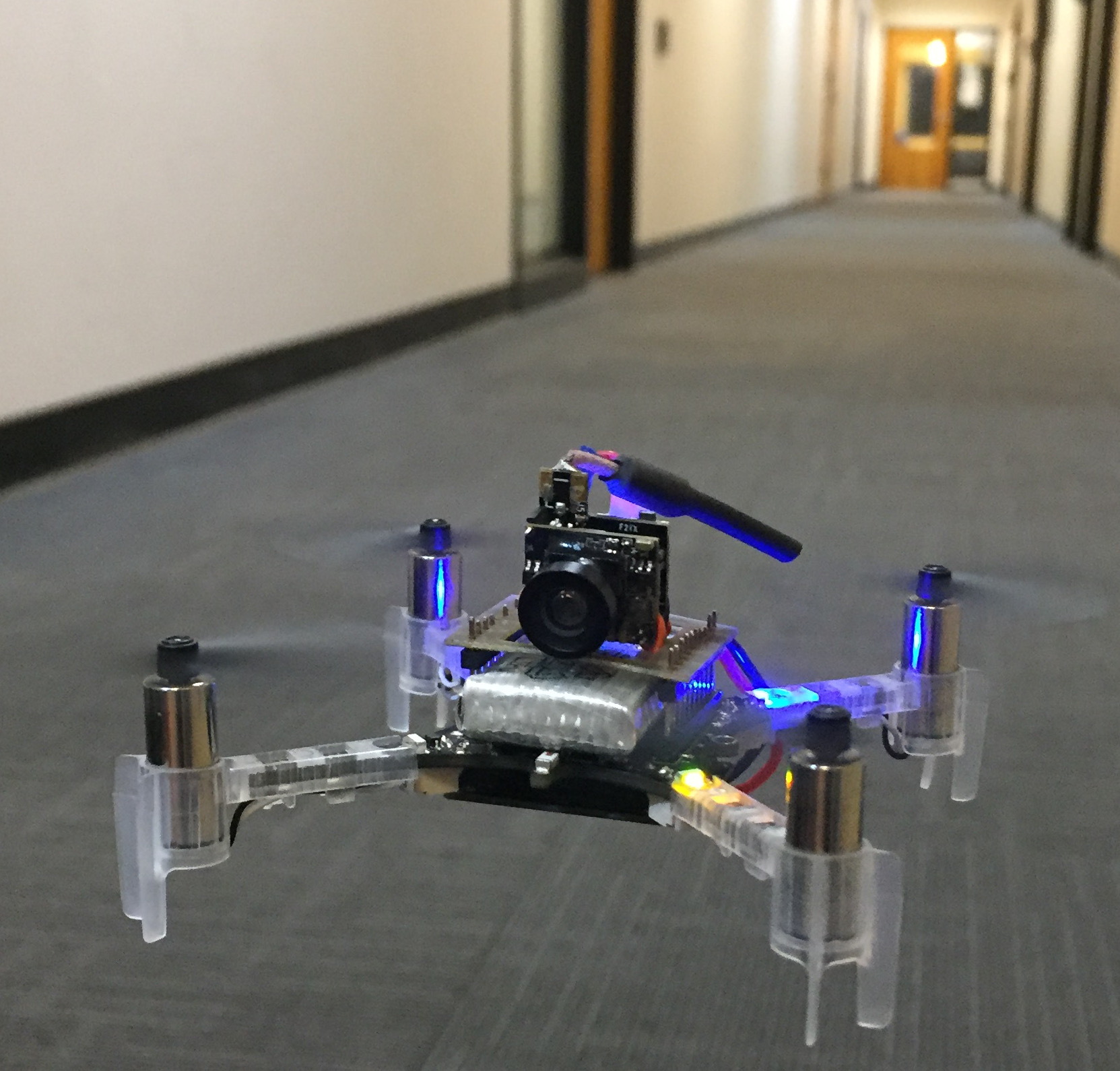}
    \caption{Our autonomous quadrotor navigating a building from raw monocular images using a learned collision avoidance policy trained with a simulator and one hour of real-world data.}
    \vspace{-0.2in}
    \label{fig:teaser}
\end{figure}

Deep reinforcement learning algorithms offer the enticing possibility of jointly automating both the perception and control systems of robots with only a minimal amount of manual engineering and a high degree of generality~\cite{li2017deep}. For example, convolutional neural network models trained with deep reinforcement learning could be used to avoid collisions and navigate inside buildings using only low-cost, low-power cameras, making them well-suited for SWaP (size, weight, and power) constrained autonomous flight~\cite{Giusti2016_RAL,Kaufmann2018_arxiv,Palossi2018_arxiv}.
However, as with all learning-based systems, the capacity of learned policies to generalize to new situations is determined in large part by the quantity and variety of the data that is available for training. While in principle autonomous robots could gather their own data directly in the real world, generalization is so strongly dependent on dataset size and diversity that it can almost always be improved simply by adding more experience, especially for fragile and safety-critical systems such as quadrotors, for which large datasets may be difficult to collect. It is therefore highly advantageous to integrate other, more plentiful sources of data into the training process. In this work, we investigate how a combination of simulated and real-world data can enable effective generalization for collision avoidance on a real-world nano aerial vehicle (NAV), shown in Fig.~\ref{fig:teaser}, using only an onboard monocular camera.

While transferring simulated experience into the real world has received considerable attention in recent years~\cite{Tzeng2015_ICCV,Ganin2016_JMLR,Sadeghi2017_RSS,Zhang2018_arxiv}, deployment of policies trained in simulation onto real-world robots poses a major challenge: complex real-world physical and visual phenomena are difficult to simulate accurately, and the systematic differences between simulation and reality are typically impossible to eliminate. Many of the prior works in this area have focused on differences that are irrelevant to the task; nuisance factors, such as variations in visual appearance, to which the optimal policy should be \emph{invariant}. Such nuisance factors can be eliminated by regularizing for invariance. However, some aspects of the simulation, particularly in terms of the dynamics of the robot, differ from reality in systematic ways that cannot be ignored. This is especially important for small-scale aerial vehicles, where air currents, drift, and turbulence are significant. In principle, this mismatch can be addressed by fine-tuning models trained in simulation on real-world data. However, as we will show in our experiments, na\"ive fine-tuning with small, real-world datasets can result in catastrophic overfitting.

In this work, we instead aim to devise a transfer learning algorithm where the physical behavior of the vehicle is learned mostly from real-world data, while simulated experience provides for a visual perception system that generalizes to new environments. In essence, real-world experience is used to learn how to fly, while simulated experience is used to learn how to generalize. Rather than simply fine-tuning a deep neural network policy using real-world data, we separate our model into a perception and control subsystem. The perception subsystem transfers visual features from simulation, while the control subsystem is trained with real-world data. This enables our approach to transfer knowledge from simulation and generalize to new real-world environments more effectively than alternative techniques. We further evaluate several choices for training the visual system in simulation, and observe that visual features that are task-oriented, such as models trained with reinforcement learning specifically for navigation and collision avoidance, transfer substantially better than task-agnostic feature learners trained with unsupervised learning~\cite{Kingma2014_ICLR} or standard supervised pre-training techniques, such as pre-training on large image recognition datasets~\cite{Russakovsky2015_IJCV}.

The main contribution of our paper is a method for combining large amounts of simulated data with small amounts of real-world experience to train real-world collision avoidance policies for autonomous flight with deep reinforcement learning. The principle underlying our method is to learn about the physical properties of the vehicle and its dynamics in the real world, while learning visual invariances and patterns from simulation. We compare a variety of methods for learning the visual features, and find that reinforcement learning in simulation leads to the most transferable representations when compared to unsupervised and supervised alternatives. On a real-world nano aerial vehicle (NAV) collision avoidance task, our method can fly $4\times$ further compared to alternative methods, and can navigate through hallways with various lighting conditions and geometry.

%%%%%%%%%%%%%%%%%%%%%%%%%%%%%%%%%%%%%%%%%%%%%%%%%%%%%%%%%%%%%%%%%%%%%%%%%%%%%%%%
\section{Related Work}

There has been much work on transfer learning for control policies~\cite{Taylor2009_JMLR}, including from simulation to reality. Prior works have sought to transfer policies by combining simulated and real-world data, including techniques such as domain adaptation~\cite{Daftry2016_ISER,Bousmalis2017_arxiv} and feature space learning~\cite{Zhang2017_ACRA,Ghadirzadeh2017_IROS}. These methods learn task-agnostic perception models, primarily in order to avoid requiring labels in the real-world. In contrast, our approach uses a task-specific perception model---specifically, the perceptual neural network layers from a policy learned in simulation---for transfer, which we demonstrate in our experiments is crucial for success.

Other approaches have sought to improve transfer by minimizing the gap between simulation and reality, either by bringing the simulator closer to reality~\cite{Hanna2017_AAAI,Lowrey2018_SIMPAR}, making reality closer to the simulator~\cite{Xia2018_CVPR,Zhang2018_arxiv}, or randomizing visual~\cite{Sadeghi2017_RSS,Pinto2018_RSS} or physical~\cite{Mordatch2015_IROS,Rajeswaran2017_ICLR,Yu2017_RSS,Peng2018_ICRA} properties of the simulator. These approaches seek to reduce the policy overfitting to systemic or irrelevant differences between simulation and reality, while our approach is complementary in that it seeks to adapt models learned in simulation to the real-world using a limited amount of real-world data. Some prior works also acknowledge the existence of the reality gap and learn to adapt these imperfect models using real-world data~\cite{Fu2016_IROS,Rusu2017_CoRL,Rastogi2018_ICRAworkshop}. In contrast to these methods, which are either evaluated on either low-dimensional or simple dynamical systems, our approach scales to raw image inputs and can cope with the highly nonlinear dynamics of a nano aerial vehicle by learning a scalable, sample-efficient, end-to-end latent dynamics model.

The general problem of addressing differences between training and test distributions has been extensively studied in the machine learning community~\cite{Pan2010_survey}. With the recent advent of large datasets, prior work has shown that deep neural networks trained on these large datasets can enable easy transfer to new tasks via fine-tuning~\cite{Donahue2014_ICML,Sharif2014_CVPR,Yosinski2014_NIPS,Hinterstoisser2017_arxiv}. In our experiments, we attempted a similar transfer by using the perception layers from a neural network model trained on Imagenet~\cite{Russakovsky2015_IJCV} and fine-tuning; however, we found that this approach performed poorly compared to our method for the task of NAV collision avoidance.

There is extensive prior work on autonomous aerial flight, including approaches that use geometric mapping and path planning~\cite{Mohta2018_ICRA,Barry2018_JFR}, imitate an expert pilot or leverage expert labelled data~\cite{Ross2013_ICRA,Giusti2016_RAL,Kaufmann2018_arxiv}, and learn from experience using large real-world datasets~\cite{Gandhi2017_IROS,Loquercio2018_RAL}. In contrast to these works, our work focuses on developing a method for adapting to the real-world using a limited amount of real-world data, which is particularly important for the SWaP-constrained NAV platform used in this work.

%%%%%%%%%%%%%%%%%%%%%%%%%%%%%%%%%%%%%%%%%%%%%%%%%%%%%%%%%%%%%%%%%%%%%%%%%%%%%%%%
\section{Problem Formulation}

Our goal is to learn a real-world control policy by leveraging data gathered in simulation in conjunction with a limited amount of real-world data. At each time step $t$, the robot selects an action $\ba_t \in \mathcal{A}$ in state $\bs_t \in \mathcal{S}$, proceeds to the next state $\bs_{t+1}$ according to an unknown transition distribution $T(\bs_{t+1} | \bs_t, \ba_t)$, and receives a task-specific reward $r_t$. The objective of the robot is to learn the parameter vector $\theta$ of a policy distribution $\pith(\ba_t | \bs_t)$ such that the expected sum of discounted future rewards $\E_{\pith,T}[\sum_{t'=t}^{\infty} \gamma^{t'-t} r_{t'}]$ is maximized, in which the discount factor $\gamma \in [0, 1)$ determines to what degree the robot cares about rewards in the distant future.

% sim versus real
In order to train this policy $\pith$, we assume we have access to both a simulator and a small dataset collected by the robot acting in the real world.  The goal is therefore to learn a real-world policy $\pith$ using a real-world dataset $(\bs_t, \ba_t, r_t) \in \dreal$, in combination with a simulated dataset $\dsim$, such that $\pith$ generalizes well in the real world.

%%%%%%%%%%%%%%%%%%%%%%%%%%%%%%%%%%%%%%%%%%%%%%%%%%%%%%%%%%%%%%%%%%%%%%%%%%%%%%%%

\begin{figure*}[t]
    \centering
    \includegraphics[width=\linewidth]{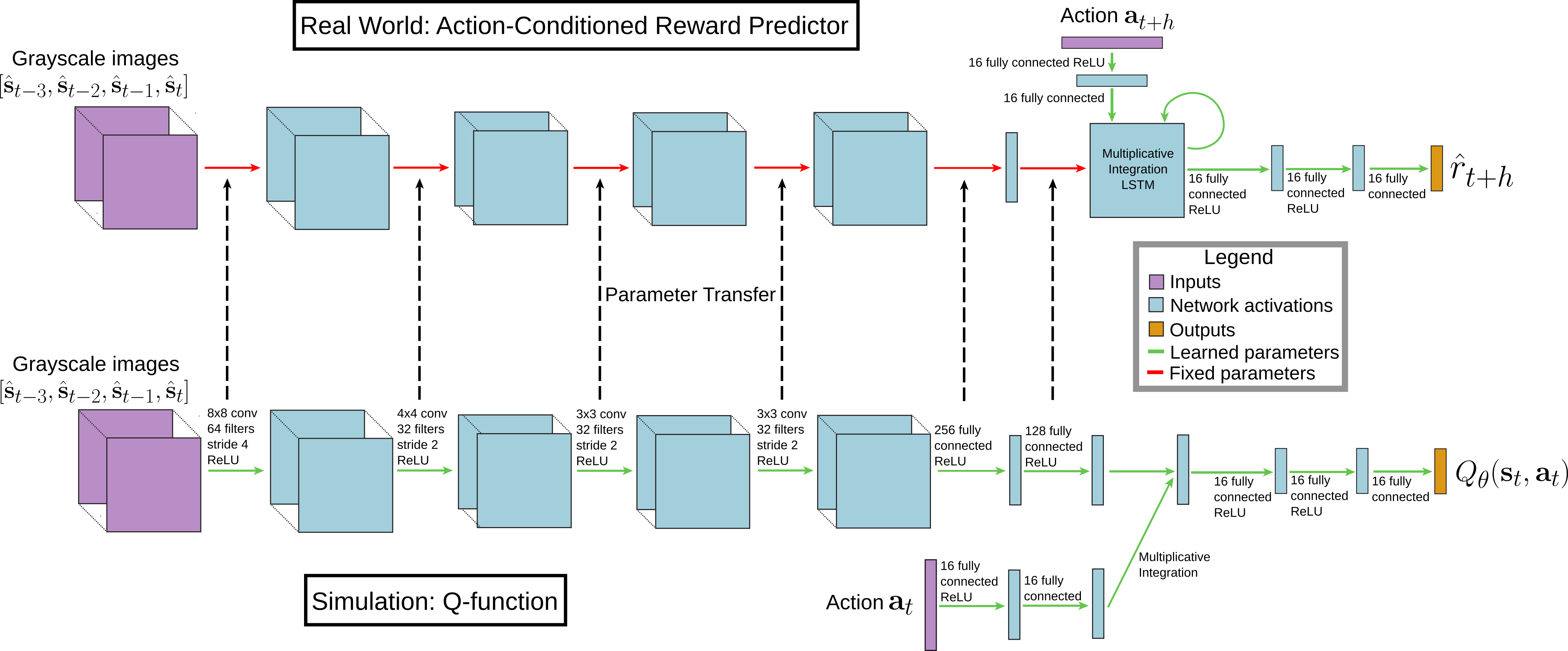}
    \caption{Our approach for leveraging both a simulator and real-world data. In simulation, we run reinforcement learning in order to learn a task-specific deep neural network Q-function model. Using real-world data from running the robot, we learn a deep neural network model that predicts future rewards given the current state and a future sequence of actions; this model can be used to form a control policy by selecting actions that maximize future rewards. In order to learn a generalizable reward prediction model with only an hours worth of real-world data, we transfer the perception neural network layers from the Q-function trained in simulation to be the perception module for the reward predictor. Our experiments demonstrate that (1) fine-tuning the Q-function on real-world data does not lead to good performance, (2) the reward predictor is better suited for real-world learning due to the limited amount of real-world data, and (3) learning a task-specific model in simulation improves transfer of the perception module.}
    \label{fig:method-nn}
    \vspace*{-18pt}
\end{figure*}

\section{Generalization through Simulation}

We will now describe our approach for real-world robot learning using generalization through simulation. Our key insight is that the real world and simulation can serve complementary functions for robot learning: data gathered in the real world provides accurate signals about the dynamics of the robot, but suffers from a lack of visual diversity due to the difficulty of gathering experience in the real-world, while simulation provides an easy way to gather large amounts of visually diverse data, but suffers from unrealistic dynamics. Our approach therefore uses the real world data for learning the dynamics of the robot, while leveraging the simulation data to learn a generalizable visual perception system. We will first describe our real-world control policy learning approach, and then discuss how to transfer a visual perception system learned in simulation to enable the real-world policy to generalize.

\subsection{Real-World Policy Learning}

Given that we will only have access to a small amount of real-world data, we require a policy learning algorithm that is sample-efficient. We therefore build off of the generalized computation graph~\cite{Kahn2018_ICRA}, a flexible policy learning framework in which the user can instantiate the generalized computation graph to suit their specific task. In this work, we will instantiate the graph as an action-conditioned reward predictor $G_\theta(\bs_t, \bA_t^H)$ that takes as input the current state $\bs_t$ and a sequence of $H$ future planned actions $\bA_t^H = (\ba_t, ..., \ba_{t+H-1})$, and outputs the predicted rewards $\hat{R}_t^H = (\hat{r}_t, ..., \hat{r}_{t+H-1})$ at each time step in the future.

At training time, the model parameters are updated using the real-world dataset to minimize the reward prediction error
\vspace*{-10pt}
\begin{align}
    \theta^* = \arg\min_\theta \sum_{(\bs_t, \bA_t^H, R_t^H) \in \dreal} \| G_\theta(\bs_t, \bA_t^H) - R_t^H \|^2, \label{eqn:method-model-train}
\end{align}
while at test time, the learned action-conditioned reward predictor is used by a finite-horizon optimal controller to select an action sequence that maximizes the predicted future rewards
\vspace*{-5pt}
\begin{align}
    \bA^* = \arg\max_{\bA} \sum_{h=0}^{H-1} \gamma^{h} \hat{r}_{t+h}. \label{eqn:finite-horizon-controller}
\end{align}
At each time step, the controller solves for the optimal action sequence by solving Eqn.~\ref{eqn:finite-horizon-controller}, executing the first action of the resulting action sequence, proceeding to the subsequent state, and then repeating this process in a receding horizon model predictive control (MPC) fashion. In order to actually find the optimal action sequence in Eqn.~\ref{eqn:finite-horizon-controller}, we resort to approximate optimization methods. In particular, we use the cross entropy method (CEM)~\cite{deng2006cross}, which is a zeroth order stochastic optimization procedure.

This action-conditioned reward prediction approach is similar to value-based methods, such as Q-learning~\cite{Watkins1992_ML}, in that both learn to predict future values. However, Q-learning used with deep neural networks typically only works with large amounts of data~\cite{Mnih2015_nature}, and can be noticeably unstable in low-data regimes due to the Bellman bootstrap update, while the action-conditioned reward prediction approach is more stable because it reduces to standard supervised learning.

The action-conditioned reward prediction approach is also similar to model-based control methods, which learn dynamics models from state transition data. However, our approach essentially learns the dynamics of the \emph{reward} as a function of the current image state and a sequence of planned future actions. The action-conditioned reward predictor is oftentimes advantageous compared to model-based methods when the robot state $\bs$ is high-dimensional---such as in this work, in which we consider the state of the NAV to be the current image because we do not have access to the underlying state information---because learning dynamics models for image sequences, while possible~\cite{Oh2015_NIPS}, is difficult because it requires predicting a complex and high-bandwidth sensory signal. Although different, this action-conditioned reward prediction approach inherits the favorable sample-efficiency property of model-based algorithms~\cite{Kober2013_IJRR} because the learning procedure (Eqn.~\ref{eqn:method-model-train}) reduces to supervised learning. The main limitation of this approach is that the action-conditioned reward predictor only predicts $H$ steps into the future and can therefore only reason about rewards within horizon $H$. However, given the short horizon nature of our target application---collision avoidance for NAVs---we found this approach to be more than sufficient.

We now instantiate the action-conditioned reward predictor as a deep neural network. Fig.~\ref{fig:method-nn} depicts the neural network architecture. The image state $\bs_t$ is provided as input to a convolutional neural network, which outputs a latent representation of the state. This latent state then serves as the initial state of a latent dynamical system module, implemented as a recurrent neural network, which updates the latent state $H$ times using the action sequence $\bA_t^H$. Each of the $H$ latent states is then passed through fully connected layers to produce the final reward predictions $\hat{R}_t^H$.

\subsection{Transferring Visual Perception Systems from Simulation}

Although the action-conditioned reward predictor is a sample-efficient policy learning algorithm, the policy is still prone to overfitting to the training data and may therefore fail to generalize to novel real-world environments due to the immense visual diversity of the real-world. We therefore seek to leverage simulation data in order to enable better real-world policy generalization.

In deciding how to leverage our simulator, we make two key observations: (1) current state-of-the-art simulators are good at providing realistic and diverse visual scenes~\cite{Xia2018_CVPR}, but do not accurately model the complex, real-world dynamics of NAVs and (2) the model learned in simulation should be task-specific and align with the real-world robot task in order for the model to learn to distill task-relevant features. Our approach will therefore learn a task-specific model in simulation, and then transfer the visual perception system part of the model to the real-world policy.

\textbf{Learning a task-specific model.} The task-specific model we learn is a deep neural network Q-function $Q_\theta(\bs_t, \ba_t)$ that is trained using Q-learning~\cite{Watkins1992_ML}; this Q-function represents the expected sum of future rewards an agent would achieve in state $\bs_t$, executing action $\ba_t$, and acting optimally thereafter. We use the Q-learning algorithm, as opposed to the action-conditioned reward predictor used for real-world policy learning, because (1) we have access to large amounts of data in simulation, which is a requirement for deep Q-learning, and (2) Q-learning can learn long-horizon tasks, which may improve the visual features that it learns.

Q-learning updates the parameters of the Q-function by minimizing the Bellman error for all state, action, reward, next state tuples in the (simulation-gathered) dataset:
\begin{align*}
    \theta^* = \arg\min_\theta \sum_{\dsim} \| Q_\theta(\bs_t, \ba_t) - [r_t + \gamma \max_{\ba'} Q_\theta(\bs_{t+1}, \ba')] \|^2 .
\end{align*}
Using the Q-function, optimal actions can then be selected by finding the action that maximizes the Q-function:
\begin{align*}
    \ba^* = \arg\max_{\ba} Q_\theta(\bs_t, \ba).
\end{align*}
In deep Q-learning algorithms with discrete action spaces, this maximization can be performed optimally. However, for deep Q-functions that take as input continuous actions, this maximization can be approximated using stochastic optimization techniques~\cite{Kahn2018_ICRA,Quillen2018_ICRA}.

The Q-function neural network model is shown in Fig.~\ref{fig:method-nn}. The model consists of three distinct neural network modules: a perception module consisting of a convolutional neural network for processing the input image state, an action module consisting of a fully connected neural network for processing the action, and a value module consisting of a fully connected neural network for combining the processed state and action to produce the resulting Q-value. We note that Q-function and action-conditioned reward predictor have a very similar structure, which illustrates the purpose of the generalized computation graph: both approaches have the same underlying mechanisms, but are trained slightly differently~\cite{Kahn2018_ICRA}.

\textbf{Visual perception system transfer.} We will use the visual perception neural network layers trained when learning the task-specific Q-function in order to transfer the visual perception system from simulation to the real-world. Concretely, we will initialize the weights of the real-world policy's visual perception layers (Fig.~\ref{fig:method-nn} top) to the values of the visual perception layers from the Q-function learned in simulation (Fig.~\ref{fig:method-nn} bottom), and hold these perception layers fixed during real-world policy training. Although these layers could be further fine-tuned using the real-world data, we decided to hold these layers fixed to prevent the real-world policy from overfitting to the training data.

\subsection{Algorithm Overview}

We now provide a brief summarizing overview of our approach. First, we train a deep neural network Q-function using deep reinforcement learning in a visually diverse set of simulated environments. Then, we create the deep neural network action-conditioned reward prediction model, in which we use the perception layers from the simulation-trained Q-function to process the input image state. Next, we train the action-conditioned reward prediction model using real-world data gathered by the robot; however, when training the model, we do not update the parameters of the perception layers. Using this action-conditioned prediction model trained on real-world data, but leveraging a task-specific visual perception system trained in simulation, our real-world policy will be better able to generalize to novel environments.

%%%%%%%%%%%%%%%%%%%%%%%%%%%%%%%%%%%%%%%%%%%%%%%%%%%%%%%%%%%%%%%%%%%%%%%%%%%%%%%%

\setlength{\tabcolsep}{3pt}
\begin{table*}
\centering
\begin{tabular}{|c|c|c|c|c|c||c|c|}
\hline & \specialcell{Simulation\\Model} & \specialcell{Perception\\System\\Transferred} & \specialcell{Real-World\\Learned Model} & \specialcell{Uses Real-\\World Data} & \specialcell{Perception Layers\\Trained with\\Real-World Data} & \specialcell{Time Until Collision\\(seconds, max 86)} & \specialcell{Percentage\\Hallway\\Traversed}\\
\hline sim only & Task-specific & N/A & N/A & \xmark & N/A & 16.5 (0.5) & 19 \\
\hline sim fine-tuned & Task-specific & \xmark & Q-function & \cmark & \cmark & 6.0 (28.5) & 7 \\
\hline sim fine-tuned perception fixed & Task-specific & \xmark & Q-function & \cmark & \xmark & 6.5 (66.5) & 8 \\
\hline real-world only & N/A & \xmark & ACRP & \cmark & \cmark & 7.8 (30.0) & 9 \\
\hline supervised (ImageNet) transfer & N/A & \cmark & ACRP & \cmark & \xmark & 9.5 (4.5) & 11 \\
\hline unsupervised (VAE) transfer & Task-agnostic & \cmark & ACRP & \cmark & \xmark & 21.0 (19.3) & 24 \\
\hline \textbf{GtS (ours)} & Task-specific & \cmark & ACRP & \cmark & \xmark & \textbf{85.8 (2.5)} & \textbf{100} \\
\hline
\end{tabular}
\caption{Comparison of our generalization through simulation (GtS) approach with prior methods for the task of flying down a straight hallway. Note that this hallway was not in the real-world training data. Each approach and baseline can be characterized with five properties: (1) is the simulation model used for transfer task-specific or task-agnostic? (2) is the real-world perception module transferred from another model? (3) is the model trained in the real world a neural network Q-function or a neural network action-conditioned reward predictor? (4) is real-world data used for training? and (5) are the perception layers trained with the real-world data or held fixed? Each approach attempted the task 5 times, and the time to collision (median and interquartile range) was recorded. Our approach was able to reliably fly down the entire hallway without colliding, consistently reaching the maximum flight time.}
\label{table:exp-results}
\vspace*{-7pt}
\end{table*}

\section{Experiments}

We evaluate our approach on a collision avoidance task with a nano aerial vehicle (NAV). This platform is well-suited for testing our transfer learning approach because it is SWaP constrained. The NAV we use is the Crazyflie 2.0 nano quadcopter~\cite{bitcraze2016crazyflie}, shown in Fig.~\ref{fig:teaser}. The Crazyflie has dimensions 92x92x29mm and weighs 27 grams. The action space consists of forward speed, yaw rate, and height, which is enabled by a downward-facing optical flow and height sensor. To allow for perceptual navigation, we added a 3.4 gram monocular camera to the Crazyflie. With the added weight, the maximum flight duration is approximately four minutes. Communication with the Crazyflie is done via a radio-to-USB dongle connected to a nearby laptop. All action selection using the learned policies is performed on this laptop, but could be deployed on the NAV in future work~\cite{Palossi2018_arxiv}.

\begin{wrapfigure}{r}{0.5\columnwidth}
    \vspace*{-10pt}
	\centering
	% left bottom right up
	\includegraphics[height=0.056\textheight,trim={0.45cm 0.7cm 0.8cm 0.5cm},clip]{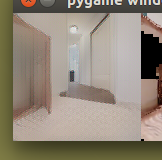}
	\hfill
	\includegraphics[height=0.056\textheight,trim={0.45cm 1.2cm 0.8cm 0.5cm},clip]{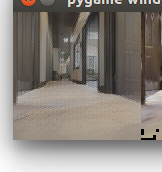}
	\hfill
	\includegraphics[height=0.056\textheight,trim={0.15cm 0.2cm 0.35cm 0.3cm},clip]{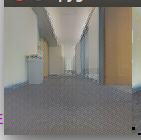}
	\caption{A subset of the environments used for simulation training.}
	\label{fig:exp-gibson}
	\vspace*{-10pt}
\end{wrapfigure}

% describe simulation
For training the simulation policy, we used the Gibson simulator~\cite{Xia2018_CVPR}, which contains a large variety of 3D scanned environments (Fig.~\ref{fig:exp-gibson}). We modelled the quadrotor as a camera with simple point mass dynamics, meaning that the actions directly control the pose of the robot camera. Although these dynamics are a severe oversimplification of real-world NAV dynamics, the goal of the simulator is not to accurately simulate the NAV, but rather to enable the collection of a visually diverse set of data that can then be used to train a task-specific model for the purpose of visual transfer. We will show in our experiments that even with this oversimplified dynamics model, we are still able to successfully transfer the visual perception system from our simulation-trained model.

Simulation data was gathered by running separate instances of Q-learning in 16 different environments. The reward function for Q-learning was 0 for no collision, and -1 for collision. After all the instances of Q-learning finished training, we trained a single Q-function on all of the 17 million simulation-gathered data points. Real-world data was gathered by running the simulation-trained policy in a single hallway on the 5th floor of Cory Hall at UC Berkeley for one hour, resulting in 14,000 data points.

For both simulation and the real world, the state consisted of the four most recent camera image converted to grayscale and downsized to a resolution of $72 \times 96$, resulting in the state space $\mathcal{S} \in \mathbb{R}^{4 \times 72 \times 96}$. The action consisted solely of the yaw angular velocity $\mathcal{A} \in \mathbb{R}^1$ because the height and speed were held constant at 0.4 meters and 0.3 m/s, respectively. Data was gathered at 4 Hz, the discount factor $\gamma$ was set to 1, and the action-conditioned reward prediction model had a horizon of $H=12$, which corresponds to predicting 3 seconds into the future.

\newcommand{\Qrw}{\textbf{Q1}}
\newcommand{\Qacrp}{\textbf{Q2}}
\newcommand{\Qtaskspec}{\textbf{Q3}}
\newcommand{\Qperception}{\textbf{Q4}}
In our experimental evaluation, we seek to answer the following questions:
\begin{enumerate}
    \item[\Qrw] Does including real-world data improve performance?
    \item[\Qacrp] Does the action-conditioned reward predictor lead to better real-world policies compared to Q-learning?
    \item[\Qtaskspec] Is a task-specific or task-agnostic simulation-trained model better for real-world transfer?
    \item[\Qperception] Does transferring the perception module from the simulation-trained model improve real-world performance?
\end{enumerate}
We compare our approach to the following methods:
\begin{itemize}
    \item[-] Sim only: The Q-function policy trained on all of the simulated data.
    \item[-] Sim fine-tuned: The Q-function policy trained on all of the simulated data, and then fine-tuned solely on the real-world data.
    \item[-] Sim fine-tuned perception fixed: The Q-function policy trained on all of the simulated data, and then fine-tune only the non-perception layers on the real-world data.
    \item[-] Real-world only: The action-conditioned reward predictor trained solely on the real-world data.
    \item[-] Supervised (ImageNet) transfer: Using pre-trained convolutional features from a model~\cite{Howard2017_arxiv} trained on Imagenet~\cite{Russakovsky2015_IJCV} for the perception module, and training the action-conditioned reward predictor using the real-world data with the perception layers held fixed.
    \item[-] Unsupervised (VAE) transfer: A variational autoencoder~\cite{Kingma2014_ICLR} generative model is trained on the simulated image data. The encoder, which maps input images to a concise latent state, is then used as the perception module for the action-conditioned reward predictor, which is trained on the real-world data.
\end{itemize}

\begin{figure}
    \captionsetup[subfigure]{aboveskip=1pt,belowskip=-7pt}
	% left bottom right up
	\begin{subfigure}[t]{\columnwidth}
	    \caption{Straight Hallway}
		\includegraphics[width=0.32\columnwidth,trim={5cm 6.5cm 5cm 7.5cm},clip]{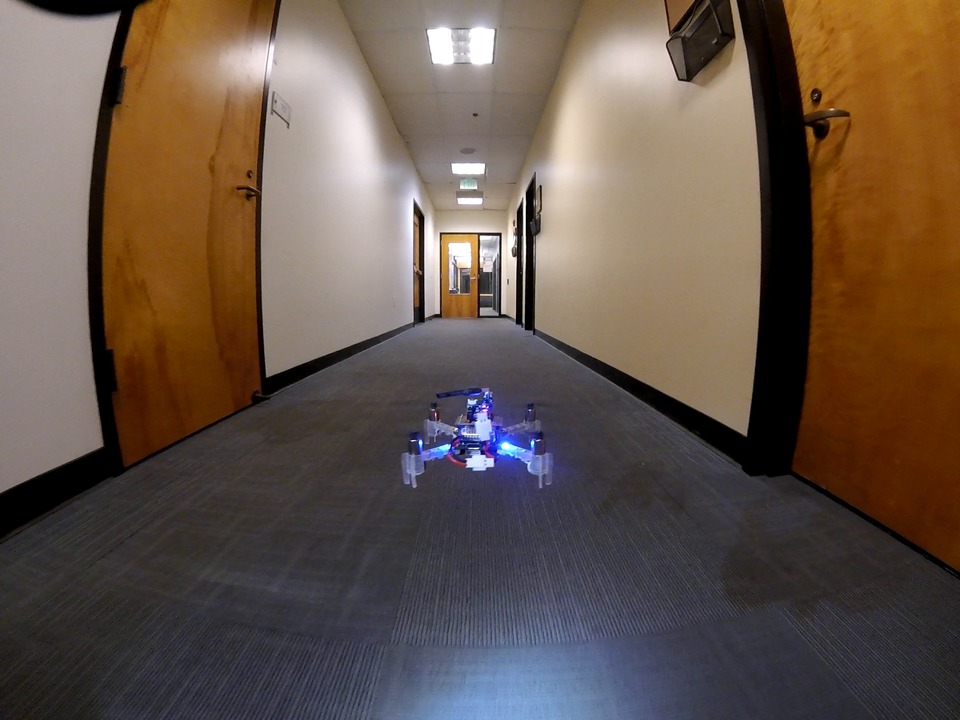}
% 		\hfill
		\includegraphics[width=0.32\columnwidth,trim={5cm 6.5cm 5cm 7.5cm},clip]{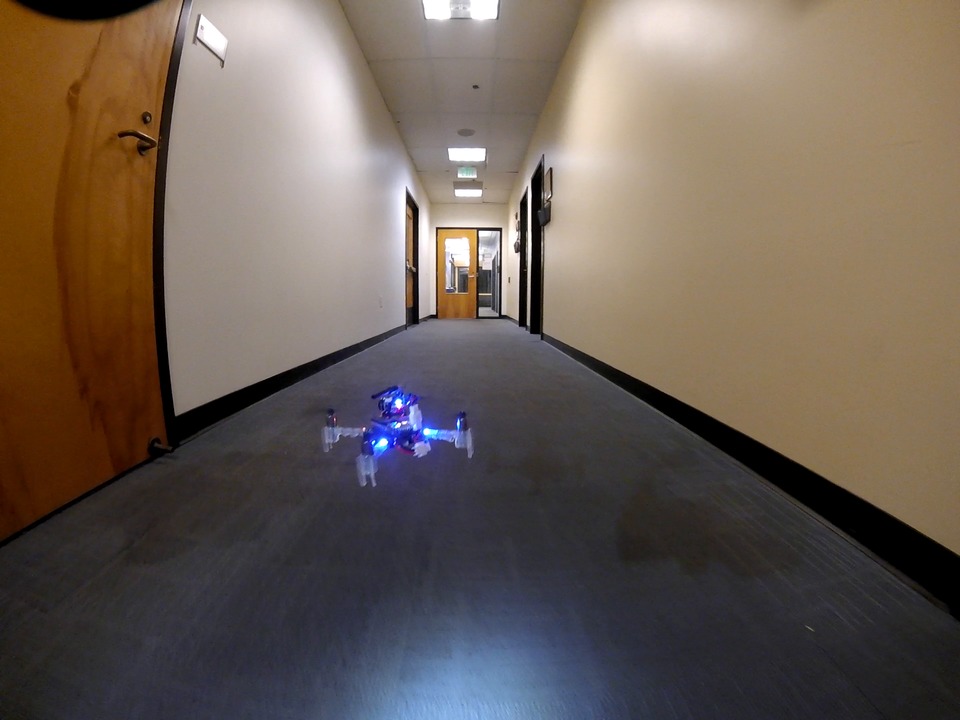}
% 		\hfill
		\includegraphics[width=0.32\columnwidth,trim={5cm 6.5cm 5cm 7.5cm},clip]{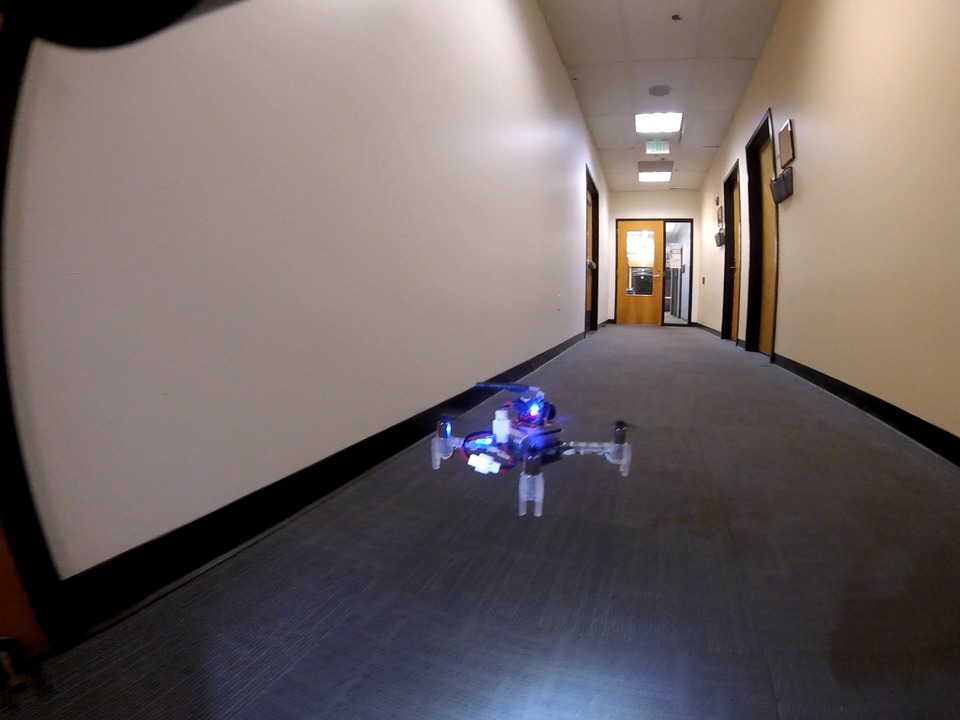}
		\label{fig:exp-crazyflie-straight}
	\end{subfigure}
	\begin{subfigure}[t]{\columnwidth}
		\includegraphics[width=0.32\columnwidth,trim={0cm 0cm 0cm 0cm},clip]{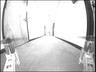}
% 		\hfill
		\includegraphics[width=0.32\columnwidth,trim={0cm 0cm 0cm 0cm},clip]{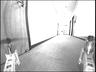}
% 		\hfill
		\includegraphics[width=0.32\columnwidth,trim={0cm 0cm 0cm 0cm},clip]{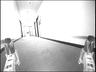}
	\end{subfigure}
	\begin{subfigure}[t]{\linewidth}
	    \caption{Curved Hallway}
		\includegraphics[width=0.32\columnwidth,trim={0cm 0cm 0cm 0cm},clip]{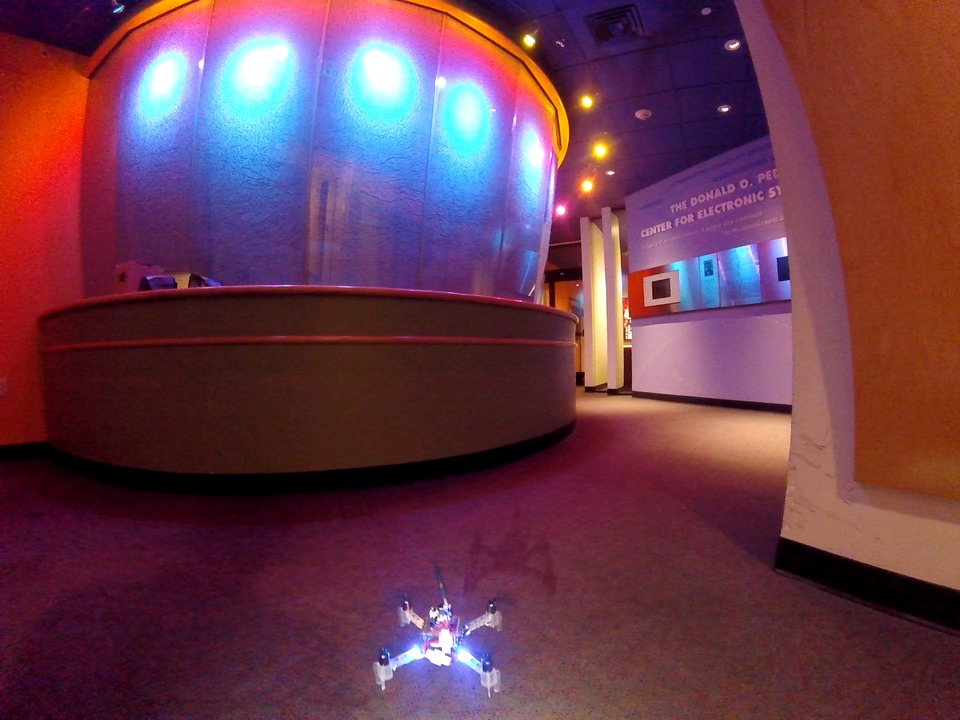}
% 		\hfill
		\includegraphics[width=0.32\columnwidth,trim={0cm 0cm 0cm 0cm},clip]{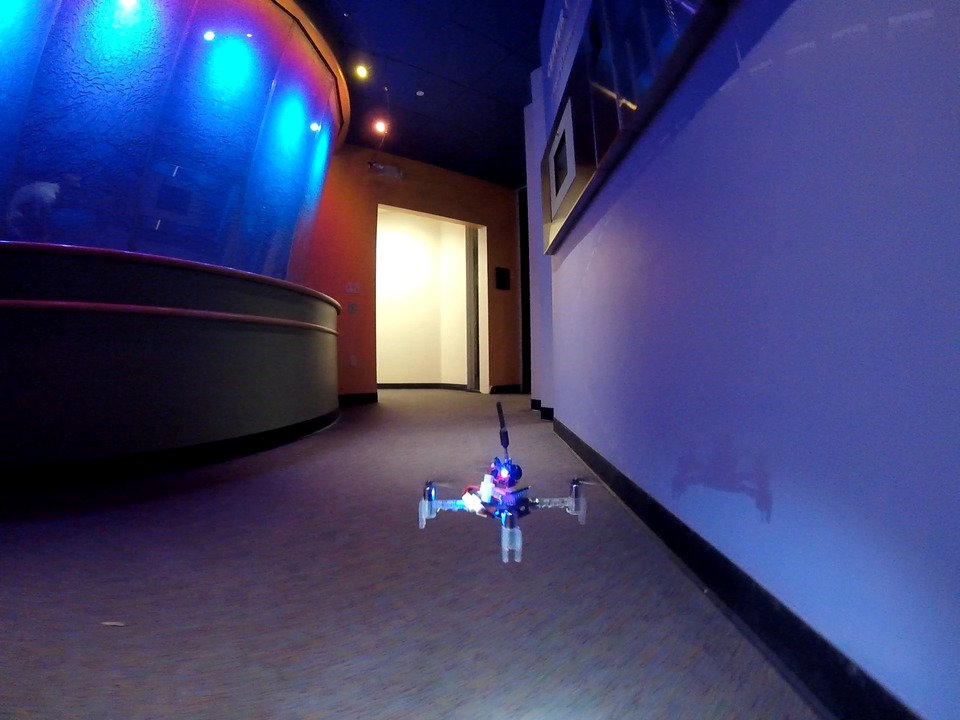}
% 		\hfill
		\includegraphics[width=0.32\columnwidth,trim={0cm 0cm 0cm 0cm},clip]{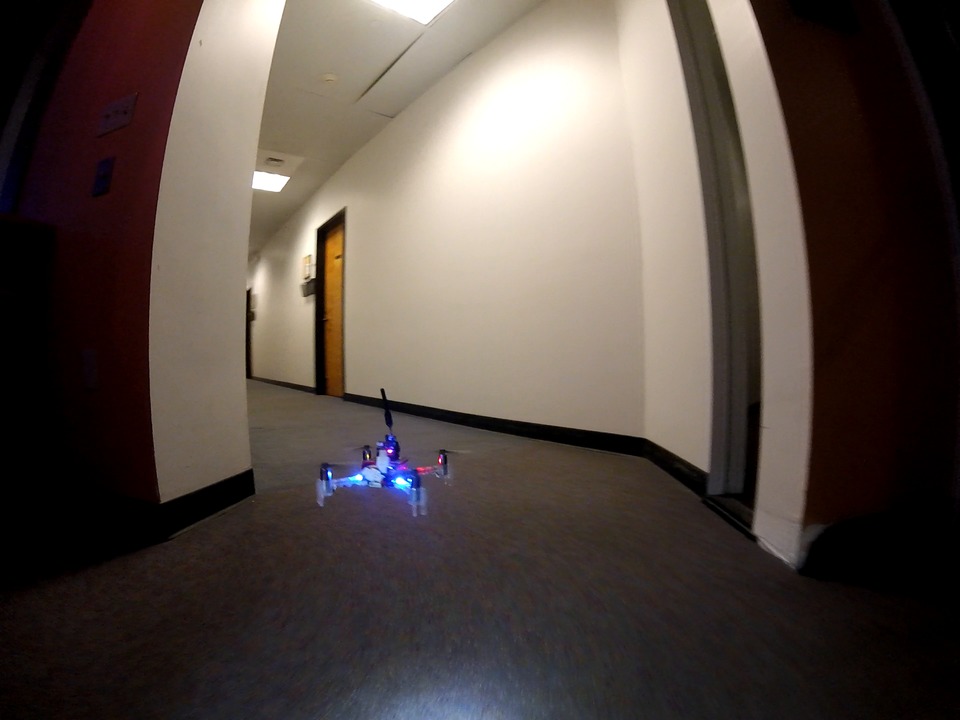}
		\label{fig:exp-crazyflie-curve}
	\end{subfigure}
	\begin{subfigure}[t]{\linewidth}
		\includegraphics[width=0.32\columnwidth,trim={0cm 0cm 0cm 0cm},clip]{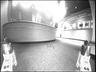}
% 		\hfill
		\includegraphics[width=0.32\columnwidth,trim={0cm 0cm 0cm 0cm},clip]{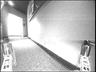}
% 		\hfill
		\includegraphics[width=0.32\columnwidth,trim={0cm 0cm 0cm 0cm},clip]{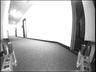}
	\end{subfigure}
	\begin{subfigure}[t]{\linewidth}
	    \caption{Zig-zag Hallway}
		\includegraphics[width=0.32\columnwidth,trim={0cm 5cm 0cm 5cm},clip]{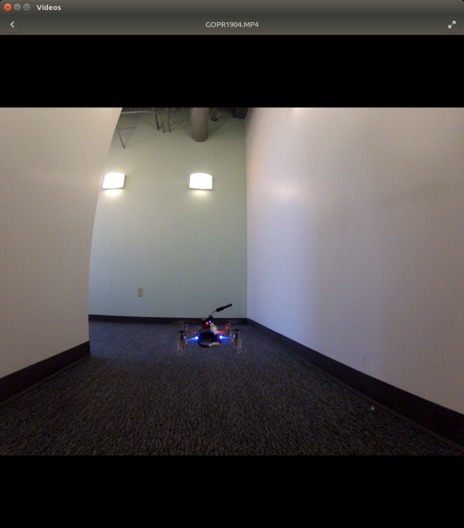}
% 		\hfill
		\includegraphics[width=0.32\columnwidth,trim={0cm 5cm 0cm 5cm},clip]{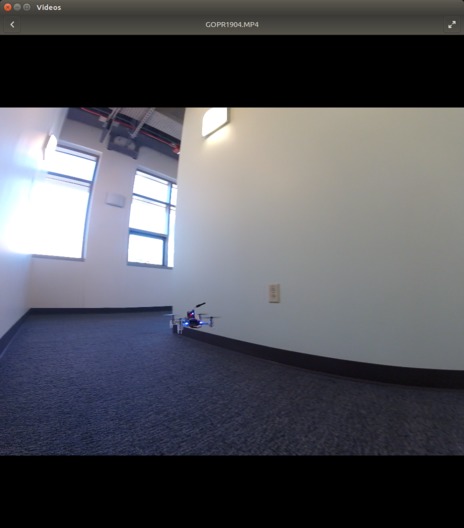}
% 		\hfill
		\includegraphics[width=0.32\columnwidth,trim={0cm 5cm 0cm 5cm},clip]{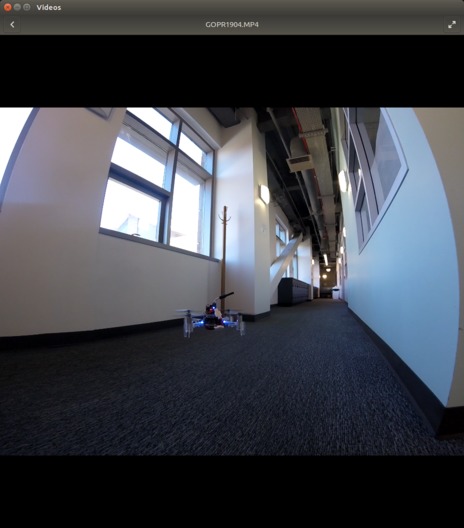}
		\label{fig:exp-crazyflie-zigzag}
	\end{subfigure}
	\begin{subfigure}[t]{\linewidth}
		\includegraphics[width=0.32\columnwidth,trim={0cm 0cm 0cm 0cm},clip]{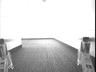}
% 		\hfill
		\includegraphics[width=0.32\columnwidth,trim={0cm 0cm 0cm 0cm},clip]{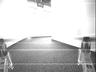}
% 		\hfill
		\includegraphics[width=0.32\columnwidth,trim={0cm 0cm 0cm 0cm},clip]{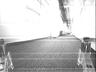}
	\end{subfigure}
	\caption{Our learning-based approach, using only the onboard, grayscale, $72 \times 96$ resolution camera images, flying through a straight, curved, and zig-zag hallway.}
	\label{fig:exp-crazyflie-all}
	\vspace*{-15pt}
\end{figure}

In order to evaluate the generalization capabilities of our approach, we present results in hallways not present in the real-world training dataset. Table~\ref{table:exp-results} compares our generalization through simulation approach with all of the considered prior methods in a novel straight hallway, and Fig.~\ref{fig:exp-crazyflie-straight} shows first- and third-person images of the NAV flying using our approach. Our method consistently flew the full length of the hallway without colliding, while the best prior method could only fly down a quarter of the hallway before colliding. Although flying down a straight hallway appears to be an easy task, the NAV drifts substantially due to imprecise sensors and environmental disturbances, and therefore avoiding collisions is non-trivial.

The sim only approach did not perform well and typically collided at doors. The sim fine-tuned and sim fine-tuned perception fixed models had trials that were indeed better than the sim only model, however their performances were inconsistent, indicating that Q-learning methods have difficulty fine-tuning on a limited amount of real-world data. Meanwhile, the policy trained solely on real-world data made some progress, but likely did not perform well due to overfitting.

In contrast, our approach, which combines learning with real-world data with simulation model pre-training, results in improved real-world performance ({\Qrw} and \Qperception). Additionally, compared to the sim fine-tuned and sim fine-tuned perception fixed approaches, our method is able to leverage both a simulation-trained model and real-world data, indicating that the action-conditioned reward predictor is crucial for sample-efficient and stable learning (\Qacrp). Lastly, our approach outperforms methods that transfer perception modules from task-agnostic models, showing that training task-specific models in simulation is beneficial for transfer (\Qtaskspec).

\begin{table}
\centering
\begin{tabular}{|c|c|c|c|}
\hline \specialcell{\% Successful Trials\\(out of 5)} & \specialcell{Straight Hallway\\with Tilted Camera} & \specialcell{Curved\\Hallway} & \specialcell{Zig-zag\\Hallway} \\
\hline Sim only                    & 0       & 0       & 0  \\
\hline Unsupervised (VAE) transfer & 0       & 0       & 0  \\
\hline \textbf{GtS (ours)}         & 80      & 60      & 80 \\
\hline
\end{tabular}
\caption{Comparison of our generalization through simulation approach with the best two prior methods from Table~\ref{table:exp-results} on three more difficult tasks. Our approach succeeds in the majority of the trials, while the prior methods fail.}
\label{table:exp-results-additional}
\vspace*{-15pt}
\end{table}

\begin{wrapfigure}{r}{0.4\columnwidth}
	\vspace*{-10pt}
	\includegraphics[width=0.4\columnwidth]{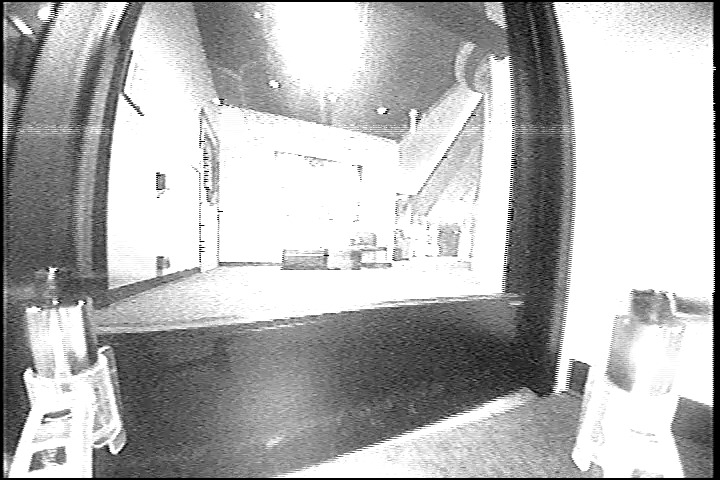}
	\caption{Example failure: collision with a glass door.}
	\label{fig:exp-glass-fail}
	\vspace*{-20pt}
\end{wrapfigure}

We also ran three additional experiments comparing our approach to the two best approaches in the straight hallway---sim only and VAE transfer---in the same straight hallway, but with the camera angle tilted down by 20 degrees, a curved hallway with varying lighting, and a zig-zag hallway. Table~\ref{table:exp-results-additional} summarizes the results, and Fig.~\ref{fig:exp-crazyflie-all} shows first- and third-person images of our approach flying. Our approach was able to fly through these difficult environments the majority of the trials, while the best prior methods were entirely unsuccessful. When our approach did fail, it was oftentimes reasonable; for example, in the curved hallways, in 30\% of the trials the NAV collided with a glass door (Fig.~\ref{fig:exp-glass-fail}). This is not surprising: the real-world data never included glass 
doors, and furthermore, many glass doors in simulation were actually traversable in simulation. However, with additional real-world data, the NAV would hopefully learn from these errors, which is the foundation of learning-based approaches.

%%%%%%%%%%%%%%%%%%%%%%%%%%%%%%%%%%%%%%%%%%%%%%%%%%%%%%%%%%%%%%%%%%%%%%%%%%%%%%%%

\section{Discussion}

We have presented an approach for learning generalizable real-world control policies using a simulator and a limited amount of real-world data. Our generalization through simulation approach uses the simulator to learn a task-specific model, and then transfers the perception layers to a sample-efficient, action-conditioned reward predictor that is trained on real-world data. Our experiments evaluate the design decisions of our method and show that our approach enables a nano aerial vehicle to fly through novel, complex hallway environments.

The key idea behind our generalization through simulation approach is to use simulation to learn how to generalize, while using the real world data to adapt the simulated model to the dynamics of the real world. Our approach currently treats these two steps as distinct; an interesting avenue for future work is to tightly integrate simulated and real-world data collection, thus enabling the robot to intelligently leverage the simulator and the real-world in an active data collection loop. Further investigation in this area is a promising approach for real-world robot learning, particularly for vision-based autonomous flight.

%%%%%%%%%%%%%%%%%%%%%%%%%%%%%%%%%%%%%%%%%%%%%%%%%%%%%%%%%%%%%%%%%%%%%%%%%%%%%%%%

\section{Acknowledgements}

We thank the Gibson~\cite{Xia2018_CVPR} team for technical support. This work was supported by the National Science Foundation under IIS-1614653, ARL DCIST CRA W911NF-17-2-0181, DARPA, NVIDIA, Google, computing support from Amazon, and Berkeley DeepDrive.

%%%%%%%%%%%%%%%%%%%%%%%%%%%%%%%%%%%%%%%%%%%%%%%%%%%%%%%%%%%%%%%%%%%%%%%%%%%%%%%%

\bibliographystyle{IEEEtran}
\bibliography{2019_ICRA_crazyflie}

\end{document}